\documentclass[10pt,twocolumn,letterpaper]{article}

\usepackage{iccv}
\usepackage{times}
\usepackage{epsfig}
\usepackage{graphicx}
\usepackage{amsmath}
\usepackage{amssymb}
\usepackage{multirow}
\usepackage[normalem]{ulem}
\useunder{\uline}{\ul}{}
\usepackage{nicefrac}
\usepackage{balance}
\usepackage{wrapfig}
\usepackage[accsupp]{axessibility} 

\usepackage[pagebackref=true,breaklinks=true,letterpaper=true,colorlinks,bookmarks=false]{hyperref}

 \iccvfinalcopy 

\def\iccvPaperID{12440} 

\ificcvfinal\fi

\begin{document}


\title{LightDepth: Single-View Depth Self-Supervision from  Illumination Decline} 
\author{Javier Rodríguez-Puigvert$^{1}$\thanks{\textit{equal contribution}} ~~~~~~Víctor M. Batlle$^{1}$\footnote[1], ~~~~~~J.M.M. Montiel$^1$ ~~~~~~Ruben Martinez-Cantin$^1$\\
Pascal Fua$^2$ ~~~~~~Juan D. Tardós$^1$ ~~~~~~Javier Civera$^1$\\
\\
\normalsize{$^1$I3A - Universidad de Zaragoza} ~~~~~~ ~~~~~~
\normalsize{$^2$École Polytechnique Fédérale de Lausanne}\\
}

\maketitle
\ificcvfinal\thispagestyle{empty}\fi




\begin{abstract}

Single-view depth estimation can be remarkably effective if there is enough ground-truth depth data for supervised training. However, there are scenarios, especially in medicine in the case of endoscopies, where such data cannot be obtained. In such cases, multi-view self-supervision and synthetic-to-real transfer serve as alternative approaches, however, with a considerable performance reduction in comparison to supervised case.
 Instead, we propose a single-view self-supervised method that achieves a performance similar to the supervised case. In some medical devices, such as endoscopes, the camera and light sources are co-located at a small distance from the target surfaces. Thus, we can exploit that, for any given albedo and surface orientation, pixel brightness is inversely proportional to the square of the distance to the surface, providing a strong single-view self-supervisory signal. In our experiments, our self-supervised models deliver accuracies comparable to those of fully supervised ones, while being applicable without depth ground-truth data.
 
\end{abstract}
\section{Introduction}

Minimally invasive medical procedures such as gastroscopies, colonoscopies and bronchoscopies rely on endoscopes that should be as small as possible. As a result, they usually house a single camera and several light points, but neither depth nor stereo cameras. 3D reconstruction is relevant in endoscopies, as it may unlock several functionalities such as the accurate estimation of the size and shape of tumors. However, both single- and multi-view depth estimation methods present significant challenges in this domain. The lack of sufficient depth annotated data hinders the use of supervised depth learning. The presence of fluids that either obscure the view or generate specularities, the sudden illumination changes, the paucity of texture and the surface deformations hamper multi-view methods both for self-supervising deep networks and for geometry estimation. Real in-body textures and fluids are hard to simulate realistically, and the synthetic-to-real gap may be large. 



\begin{figure}[!t]
  \includegraphics[width=\linewidth]{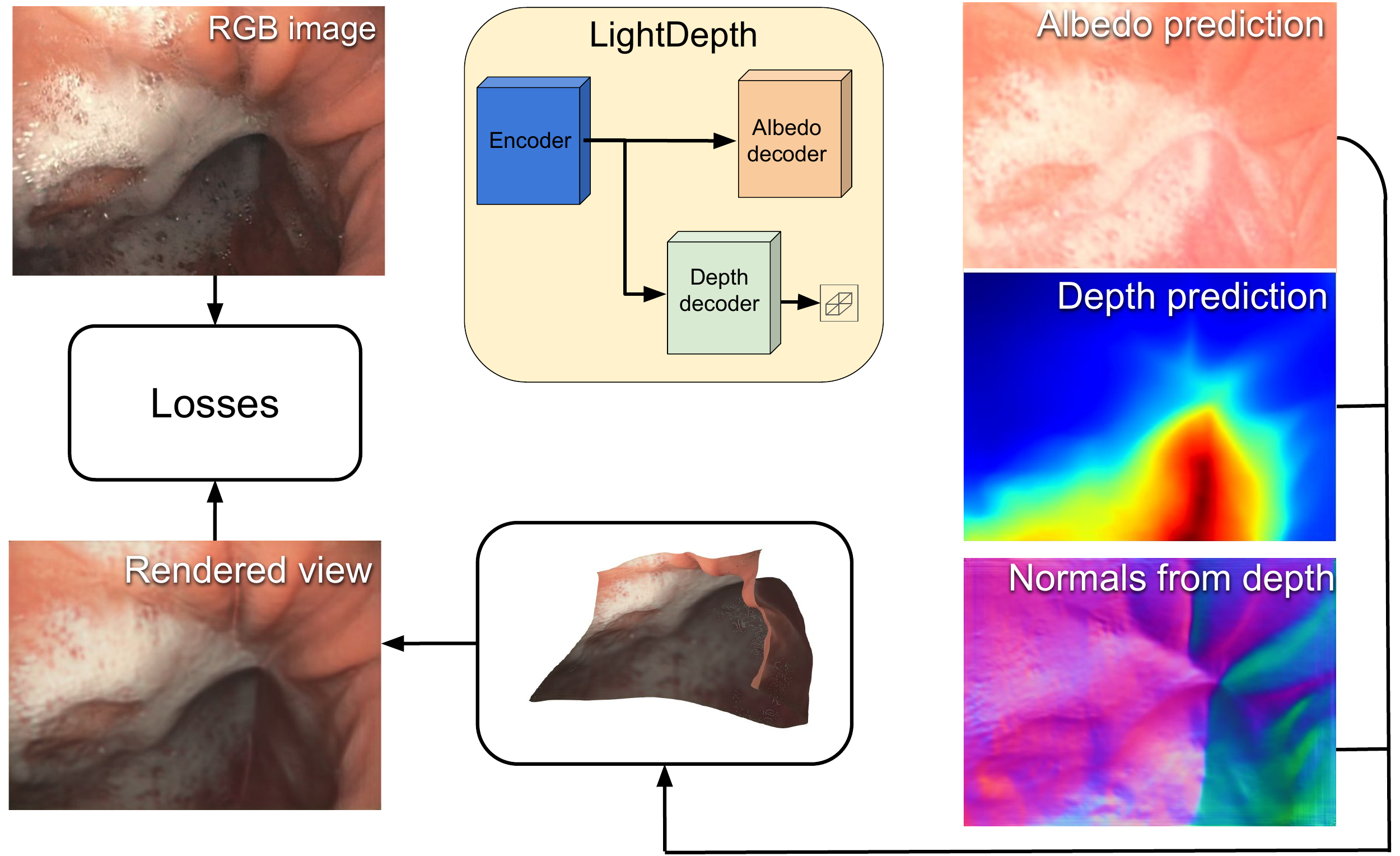}
\caption{{\bf Single-view depth self-supervision in LightDepth.} A two-headed deep network predicts albedo and depth from a single image and estimates surface normals from predicted depths. These are used to render a new image, that takes into account illumination decline and the endoscope's photometric calibration, and can be compared to the original one. Minimizing the difference between the original and rendered images is used at training time to compute the network weights and at inference time to refine the depth predictions.}
\label{fig:teaser}
\end{figure}

In this work, we propose a novel approach to depth in endoscopies that overcomes all the above challenges related to depth supervision, multi-view estimation and synthetic-to-real gaps. Our key insight is that, by exploiting a key property of endoscopic imagery, we can provide strong depth self-supervision signals from just one view. In endoscopes, the light source is rigidly located next to the camera and is close to the surface to be reconstructed. As a result, unlike in traditional shape-from-shading (SfS), points with the same albedo are imaged darker the further they are, being the decrease of intensity a function of the square distance to the light source. To exploit this, we introduce a deep network, as depicted by Figure~\ref{fig:teaser}, that estimates depths and albedos from the image, infers normals from depths, and then renders an image while taking into account the attenuation factor due to the distance between the light source and the surface. At training time, we minimize the difference between the original and rendered images. This enforces consistency of the depths, normals, and albedos and provides the required self-supervision without depth annotations. At inference, we use our trained network to predict depth from a RGB image and then, as our method is totally self-supervised, we can perform test-time refinement (TTR) for every monocular image, minimizing the difference between the input and rendered views, further refining the predicted depths.
Our quantitative evaluation on a phantom colon dataset, where ground-truth is available, shows that our {\it self-supervised} approach delivers results that are very close to that of the best supervised one, and significantly superior to that of multi-view self-supervision and synthetic-to-real transfer methods. Crucially, we show quantitatively that our method keeps working on real data, for which there is no ground-truth data that can be used for training and self-supervised alternatives underperform. 
The main specific contributions that led to such results are 1) the inclusion of illumination decline and the endoscope's photometric calibration in the rendering equation, which provides a strong supervisory signal, and 2) a single-view self-supervised method using such renders, including two-headed network architectures LightDepth U-Net and LightDepth DPT (see details in Figure \ref{fig:network}) that can be trained in large colonoscopy datasets without requiring ground truth labels and even further refined online in the test views.

\section{Related Work}

\textbf{Generic Single-view Depth Estimation.} 
It has enjoyed a renaissance after the seminal work by Eigen et al.~\cite{eigen2014}, which demonstrated the effectiveness of deep neural networks for supervised pixel-wise depth regression in natural images. Subsequent research efforts have made contributions in many different directions. To name a few, network architectures evolved to fully convolutional in Laina et al.~\cite{laina2016deeper} and more recently to transformers~\cite{Ranftl_2021_ICCV,bhat2021adabins,li2022binsformer}. Some of those works~\cite{bhat2021adabins,li2022binsformer} also discretize the continuous depth space into bins and formulate the problem as an ordinal regression, as in Fu et al. ~\cite{fu2018deep}. Other advances include interpretability~\cite{Dijk_2019_ICCV}, uncertainty quantification~\cite{poggi2020uncertainty,rodriguez2022bayesian}, and modeling camera intrinsics~\cite{facil2019cam,gordon2019depth}. 
All these approaches are supervised and require depth ground-truth data, which can be difficult and expensive to acquire. 

Self-supervised methods seek to overcome this limitation and reduce the need for ground-truth data, often by exploiting multi-view photometric consistency~\cite{monodepth17,Zhou2017,Zhou2018,Yang2018,johnston2020self,monodepth2}. This also enables depth refinement at test time~\cite{chen2019selfsup,tiwari2020pseudo,luo2020consistent,shu2020feature,watson2021temporal,izquierdo2022sfm}. Unfortunately, this kind of supervision can be noisy,  due to inaccuracies in the camera motion estimation, perspective distortions, occlusions or non-Lambertian effects, among others. As result, state-of-the-art self-supervised methods typically suffer from significantly larger inaccuracies than supervised ones. By contrast, our approach avoids these sources of errors and delivers accuracies that are close to those of supervised techniques.

\textbf{Endoscopic Single-view Depth Estimation}. 
Single-view depth estimation has been extensively studied for endoscopic purposes. Visentini et al. \cite{Visentini-Scarzanella2017} used CT renderings for depth supervision in bronchoscopies. However, CT scans in particular and ground-truth depth data in general are very rare in endoscopy, which makes self-supervision a quasi necessity. Many works explore multi-view integration \cite{luo2019details,xu2019unsupervised,huang2021self} combined with tracking and SLAM pipelines \cite{recasens2021endo,ozyoruk2021endoslam,ma2021rnnslam}. Others propose video-based training schemes~\cite{karaoglu2021adversarial,freedman2020detecting,hwang2021unsupervised}. Unfortunately multi-view self-supervision  is even more challenging  in endoscopy than in other areas due to the presence of deformations and weak texture. 

Due to the specificity of the domain, synthetic to real transfer has also been extensively explored. For example, in~\cite{shen2019context} a conditional GAN is used for depth recovery while integrating SLAM and multi-view inputs. In~\cite{chen2019slam},  a depth network is trained with synthetic images of a simple colon model and fine-tuned with domain-randomized photorealistic images rendered from CT scans. Many other works address the domain shift between simulated and real colons~\cite{mahmood2018unsupervised,mahmood2018deep,rau2019implicit,karaoglu2021adversarial,Cheng2021,Rodriguez2022}. Learning in supervised and transferring the knowledge using uncertainty~\cite{Lui2020} uses monocular videos and multi-view stereo to provide weak depth supervision. 
We will show in the results section that our approach yields more accurate results, especially given that our approach to self-supervision allows further refinement of the estimates at inference time.






\textbf{Shape from Shading (SfS).}
Depth estimation from a single image can be traced back to the early SfS methods summarized in \cite{Ruo1999} and in particular to traditional shape-from-shading~\cite{Horn89}. However, these older techniques rely on strong assumptions that do not hold in endoscopic imagery: the camera and directional point light model are located at infinity; the reflectance is Lambertian; the albedo is constant, and the surfaces are smooth. 

Importantly, lights at infinity result in ill-posed problems~\cite{Prados2005}. By contrast, when the light source is co-located with the camera that is \emph{not} distant from the target surfaces, there is a $\nicefrac{1}{d^2}$ attenuation of pixel intensity with distance $d$ to the surface, which makes the problem well-posed when the albedo is assumed to be constant. Experimental validation that this still holds when the light source is translated with respect to the optical centre is provided in~\cite{ Collins2012towards,Visentini2012}, but still assuming constant and known albedo.  Photometric stereo infers depth capturing several images from the same monocular camera under lights at different locations, but requires endoscopic hardware modifications \cite{hao2020photometric,parot2013photometric,collins20123d}.

 More recently, the topic was revisited by SIRFS (Shape, Illumination, and Reflectance from Shading)~\cite{Barron2015} that model the interdependences between shape, illumination and reflectance, and introduces statistical priors on these quantities to disentangle their effects. In subsequent works~\cite{lettry2018unsupervised,li2020inverse,sang2020single,lichy2021shape,zhang2022modeling}, priors are learned by deep neural networks using supervision, synthetic-to-real or multi-view self-supervision. In contrast, our approach does not require such priors, which makes its deployment easier.



The SfS methods applied to endoscopy require an accurate geometrical and photometrical model of the camera and light source. This can be obtained with endoscope calibration \cite{modrzejewski2020light, hao2020light, azagra2022endomapper, batlle2022photometric}.

\newcommand{\bA}{\mathbf{A}}
\newcommand{\bC}{\mathbf{C}}
\newcommand{\bI}{\mathbf{I}}
\newcommand{\bD}{\mathbf{D}}
\newcommand{\bN}{\mathbf{N}}

\section{LightDepth}
\label{sec:method}

We use a self-supervised single-view approach to train a neural network to predict the albedo, depth, and normals at every pixel of an image so that the image can be resynthesized from these values.
As shown in Fig.~\ref{fig:teaser}, we exploit this property using a dual-branch network that outputs pixel-wise depths and albedos. The normals are estimated  analytically from the depths, and, together with the albedos, are used to render images that should be as close as possible to the original ones. At the heart of this approach is the fact that the renderer takes into account light decline as a function to distance to the surface. This is what provides the necessary self-supervisory signal. 

\subsection{Photometric Model}
\label{sec:photometric-model}

\begin{figure}[ht!]
  \centering
  \includegraphics[width=0.6\linewidth]{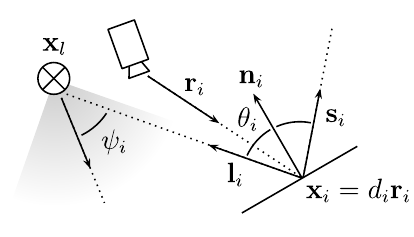}
\caption{Spotlight illumination model, a spotlight source at position $\mathbf{x}_l$ illuminates the surface point $\mathbf{x}_i$. The emission has $R(\psi_i)$ radial fall-off, suffers from an inverse-square decline with $\mathbf{x}_l \rightarrow \mathbf{x}_i$ and attenuates with the incidence angle ($\theta_i$). $\mathbf{l}_i$, $\mathbf{n}_i$, $\mathbf{r}_i$ and $\mathbf{s}_i$ are unit vectors.}
\label{fig:photo-model}
\end{figure}

As in~\cite{batlle2022photometric, modrzejewski2020light}, 
we model scene illumination  as coming from a single spotlight source located at $\mathbf{x}_l \in \mathbb{R}^3$ in the camera reference frame, as depicted by Fig.~\ref{fig:photo-model}. Spotlights usually emit with different intensities in each direction. Hence, we adopt the spotlight model (SLS) of~\cite{modrzejewski2020light}.
For surface point  $\mathbf{x}_i$ with off-axis angle $\psi_i$, we write its radiance as
\begin{align}
    \sigma_\text{SLS}(\mathbf{x}_i, \psi_i) &= \frac{\sigma_0}{\lVert \mathbf{x}_i - \mathbf{x}_l \rVert^2} R(\psi_i) \; , \\
    R(\psi_i) & = e ^ {-\mu \left(1 - \cos \left( \psi_i \right)\right)}
\end{align}
where $\sigma_0$ is the maximum radiance and $R(\psi_i)$ is the radial attenuation controlled by a spread factor $\mu$. Note that the light reaching the surface is subject to the inverse-square law and decays with the propagation distance from $\mathbf{x}_l$ to $\mathbf{x}_i$.


\textbf{Light Decline.} 
In endoscopes, the camera and the light source move jointly in a dark environment. Hence, the attenuation of the illumination is an indirect indicator of scene depth as seen from the camera. More specifically, for each pixel, we can write the rendering equation 
\begin{small}
\begin{equation}
\label{eq:im}
    \mathcal{I}(d_i, \rho_i, g)
    =
    {\left(
        \frac{\sigma_0}{\lVert d_i\mathbf{r}_i - \mathbf{x}_l \rVert^2} R(\psi_i)
        \cos\left(\theta_i \right)\;
        \rho_i \;
        g
    \right)}^{1/\gamma} \; ,
\end{equation}
\end{small}
where
$d_i$ is the depth of the $i$-th pixel with image coordinates $\mathbf{u}_i$, $\mathbf{r}_i = \pi^{-1}(\mathbf{u}_i)$ is the camera ray such that $\mathbf{x}_i = d_i\mathbf{r}_i$ and  $\pi^{-1}(\cdot)$ is the inverse projection model of the camera. $\theta_i$ stands for the light's incidence angle with respect to the surface normal $\mathbf{n}_i$, such that, $\cos\theta_i = \mathbf{l}_i \cdot \mathbf{n}_i$. $\rho_i$ represents the albedo of the surface at that point.
$g$ denotes the gain applied by the camera and $\gamma$ is the gamma correction commonly applied by cameras to adapt images to human perception. The resulting $\mathcal{I}(d_i, \rho_i, g)$ is the color captured by the camera.

Our model assumes Lambertian reflections, meaning that the light hitting the surface is scattered equally in all directions. The percentage of reflected light is known as albedo. Specular reflections, which are prevalent in endoscopic images, are not captured by this model but we will consider them in a specific loss that we describe in Section~\ref{sec:losses}.

\textbf{Calibration.} Each endoscope has different geometric and photometric parameters, the former affecting the inverse project model $\pi^{-1}$ and the latter impacting both the light position $\mathbf{x}_l$ and spread $R$.
We can estimate these parameters for a particular endoscope by minimizing the reprojection and photometric errors on images of a calibration target, similar to \cite{azagra2022endomapper, batlle2022photometric}. In our case, the auto-gain values of the endoscope are not known, so radiance measurements of the camera are unitless. Thus, we arbitrarily set $g = 1$, $\sigma_0 = 1$ and obtain up-to-scale reconstructions. Our calibration errors are between $\pm 3$ gray levels.

\begin{figure*}
  \includegraphics[width=\textwidth]{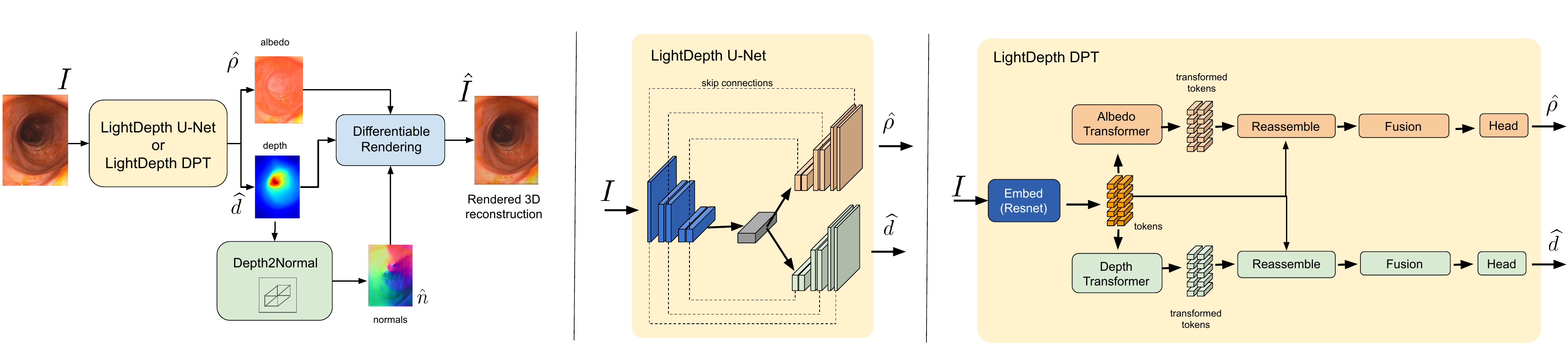}
\caption{{\bf Network Architecture.} {\bf Left.} The input image is fed into a neural network that predicts albedo and depth values for each pixel. From the estimated depths, we compute the normals at each pixel surface using a kernel-based approach. Then, the depths, albedos, and normals are sent to a differentiable renderer that takes into account illumination decline and the endoscope's photometric model, and generates a synthetic image that should be as similar as possible to the original one. We also use specular reflections in saturated pixels to self-supervise normals. We investigated two different architectures:  {\bf Center.} LightDepth U-Net is based on a standard U-Net~\cite{Ronneberger2015} with two decoding branches. {\bf Right.} LightDepth DPT is based on the DPT-Hybrid architecture~\cite{Ranftl_2021_ICCV}, with a second decoder branch added for the albedo.}
\label{fig:network}
\end{figure*}

\subsection{Self-Supervision Losses}
\label{sec:losses}

Formally, the network of Fig.~\ref{fig:teaser} takes as input an image ${I} \in [0, 1]^{w \times h \times 3}$, estimates a depth map $\widehat{d} \in (0, \infty)^{w \times h}$ and an albedo map $\widehat{\rho} \in [0, 1]^{w \times h \times 2}$ . It infers normals $\mathbf{\widehat{{n}}}$ from $\widehat{d}$, and uses $\widehat{d}$, $\mathbf{\widehat{{n}}}$, and $\rho$ to render an image $\widehat{I} \in [0, 1]^{w \times h \times 3}$  that should be as similar as possible to  ${I}$. 
To train this network, we minimize a loss
\begin{equation}
    \mathcal{L} = \mathcal{L}_{p} + \lambda_s \mathcal{L}_{s} + \lambda_{sp} \mathcal{L}_{sp} \; ,
    \label{eq:loss}
\end{equation}
where $\lambda_{s}$ and $\lambda_{sp}$ are scalar weights and $\mathcal{L}_p$,  $\mathcal{L}_s$, and $\mathcal{L}_{sp}$ are the loss terms described below.

$\mathcal{L}_p$ is a photometric loss and we take it to be the squared $L_2$  distance between the original image $I$ and the rendered one $\hat{I}$. Note that because our rendering model is fully differentiable, we can perform end-to-end training.

\begin{equation}
    \mathcal{L}_{p} = \sum_{i \in \Omega} ({I}_i -{\widehat{I}}_i)^2
    \text{,} \quad \text{ where } \quad
    \widehat{I}_i = \mathcal{I}(i, \widehat{d}_i, \widehat{\rho}_i, g)
\end{equation}

%


As in \cite{monodepth2}, $\mathcal{L}_{s}$ is a regularization term that minimizes depth gradients except in areas of high color gradients, that may correspond to depth discontinuities. We write
\begin{equation}
     \mathcal{L}_{s} =|\partial_x \widehat{d}|  e^{-|\partial_x {I}|} +|\partial_y \widehat{d}|  e^{-|\partial_y {I}|}
\end{equation}

Finally, recall that we made a Lambertian assumption in Eq.~\ref{eq:im}, which prevents us to account properly for specular reflections and the overexposed pixels they produce. This is a potential source of error and fails to exploit the very useful information that specularities provide about normals. To remedy this, we introduce specular loss $\mathcal{L}_{sp}$. Given image location $i$, the corresponding  direction $\mathbf{l}_i$ from the surface to the light source and the normal of a the surface $\mathbf{\widehat{{n}}}$, the law of reflection states that
\begin{equation}
    \mathbf{s}_i = \mathbf{l}_i - 2\mathbf{\hat{n}}_i \left( \mathbf{\hat{n}}_i \cdot \mathbf{l}_i \right)
\end{equation}
is the specularly reflected direction. Hence, we take our specular loss term to be
\begin{equation}
    \mathcal{L}_{sp} = \sum_{i\in\Omega}\left(m_i\left( \mathbf{s}_i \cdot ( - \mathbf{r}_i) - 1 \right)\right)^2~,
\end{equation}
\begin{equation}
        m_i = \left\{\begin{matrix}
        1 & I_i > th\\ 
        0 & \text{otherwise}
        \end{matrix}\right. \nonumber
\end{equation}
which minimizes the discrepancy between the expected specular reflection $\mathbf{s}_i$ and the actual direction $(-\mathbf{r}_i)$ where the camera observes the reflection, resulting in pixel with high intensity $th = 0.98$.

Our method takes a single image as input, which makes 3D shape recovery solely from pixel colors an underconstrained problem. According to Eq.~\ref{eq:im},  a change in the brightness of a pixel can be due to changes in depth, albedo, camera exposure or surface normal. For example, if a given pixel is very bright, it can be because the pixel is close to the camera/light; the surface has a different albedo, resulting in more light being reflected; the surface normal is aligned to the light/camera, which increases the reflected light; the camera exposure and digital gain have been increased, which impacts brightness values in the whole image. Given the albedo at each surface point and the camera auto-gain, we could resolve these ambiguities. However, in medical endoscopy, true albedos are unknown, and auto-gain is not provided by the hardware manufacturer. 

\paragraph{Albedo Constancy.}
We observe that endoscopy images exhibit a limited range of colors, with brighter tones being present in close areas and darker tones in deeper regions. Consequently, we hypothesize a significant correlation between albedo and the chromatic attributes, namely Hue and Saturation, in the HSV color space, as well as between depth and the Value Channel. In this way, we constrain the palette of colors that can be explained by the albedo decoder and we enhance the disentanglement between depth and albedo by setting $V=100$ for all albedo values. Hence, to predict the albedo map $\widehat{\rho}$, our network predicts just two channels per pixel, for Hue and Saturation, and assumes Value to be one to convert to the RGB space, in which the loss is formulated. 





\subsection{Network Architecture}
\label{sec:network-architecture}

Our network outputs depth and albedo maps. In Fig.~\ref{fig:network}, we provide a more detailed depiction of our encoder-decoder architecture. We have tested two different versions. The first one is a U-net with two decoders and skip connections, with a ResNet18 serving as the backbone.  Our decoders design is inspired by ~\cite{monodepth2}. The second one relies on visual transformers for depth estimation \cite{Ranftl_2021_ICCV}. As a backbone, we use a Resnet-50 (DPT-Hybrid) and two decoders that reassemble the tokens and apply attention heads. Further details regarding these architectures can be found in the supplementary material.

In both versions, to compute the normals at any given pixel, we use a convolution kernel with six-neighborhood (N, NE, E, S, SW, and W) in the depth map. We define six triangles using the central pixel as reference, with each triangle having its own normal. The normal of the central pixel is computed as the average of the normals of the triangles weighted by their area. The use of six neighbors lets us reuse triangles during the convolution pass to speed up computation.


\section{Results}\label{sec:experiments}

\begin{table*}[ht!]
    \resizebox{\linewidth}{!}{
        \begin{tabular}{clccccccccccccc}
         &  &  & \multicolumn{10}{c}{\textbf{Depth [mm]}} & \textbf{Normals [$^\circ$]} \\
        Dataset & Architecture & Backbone & Supervision &  MAE $\downarrow$ & MedAE $\downarrow$ & RMSE $\downarrow$ & RMSE\textsubscript{log} $\downarrow$ & Abs\textsubscript{Rel} $\downarrow$ & Sq\textsubscript{Rel} $\downarrow$ & $\delta < 1.25$ $\uparrow$ & $\delta < 1.25^2$ $\uparrow$ & $\delta < 1.25^3$ $\uparrow$ & MAE $\downarrow$ \\ \hline
        \multirow{2}{*}{Synthetic} & U-Net & ResNet18 & Depth GT &   \textbf{4.37} & 2.99 & \textbf{6.38} & \textbf{0.1251} & 0.0965 & \textbf{0.0008} & 0.9057 & \textbf{0.9931} & \textbf{0.9997} & 25.1 \\
         & LightDepth U-Net& ResNet18 & Light &  4.76 & \textbf{2.47} & 8.60 & 0.1375 & \textbf{0.0903} & 0.0011 & \textbf{0.9180} & 0.9820 & 0.9935 & \textbf{15.2} \\ \hline
        \multirow{9}{*}{C3VD} & U-Net& ResNet18 & Depth GT & 4.15 & 3.29 & 5.52 & 0.1139 & 0.0902 & 0.0007 & 0.9172 & {\ul0.9943} & \textbf{0.9994} & 26.5 \\
         & DPT-Hybrid \cite{Ranftl_2021_ICCV}& ResNet50 & Depth GT  & \textbf{3.22} & 2.77 & \textbf{4.10} & \textbf{0.0860} & \textbf{0.0699} & \textbf{0.0004} & \textbf{0.9640} & 0.9865 & 0.9913 & \textbf{15.1} \\
         & Monodepth2 \cite{monodepth2} &ResNet50& Multi-View & 14.27 & 9.59 & 18.64 & 0.3921 & 0.2971 & 0.0070 & 0.4897 & 0.7313 & 0.8611 & 43.6 \\
         & CADepth \cite{yan2021channel} & ResNet18 & Multi-View & 52.35 & 17.04 & 87.43 & 0.9144 & 1.1916 & 0.2650 & 0.3664 & 0.5653 & 0.6679 & 67.2 \\
          & XDCycleGAN \cite{mathew2020augmenting} & ResNet &  Cycle & 17.16 & 11.91 & 22.43 & 0.4953 & 0.3616 & 0.0105 & 0.4291 & 0.6615 & 0.7910 & 64.4 \\
         & LightDepth U-Net  & ResNet18 & Light & 4.37 & 2.92 & 6.31 & 0.1183 & 0.0856 & 0.0007 & 0.9315 & 0.9934 & \textbf{0.9994} & 24.0 \\
         & LightDepth DPT & ResNet50 &Light&3.94 &2.67 &5.60&0.1080&0.08046&0.0006&0.9476&0.9965&	\textbf{0.9994} &{ \ul 21.3}\\
        
         & LightDepth U-Net  & ResNet18& Light (TTR) & 3.72 & {\ul 2.59} & 5.43 & {\ul 0.1060} & {\ul 0.0770} & {\ul 0.0005} & {\ul 0.9505} & \textbf{0.9971} & \textbf{0.9994} &  23.5 \\
         & LightDepth DPT & ResNet50 & Light (TTR) & {\ul 3.70}&\textbf{2.58}& {\ul 5.27}	&0.1073&	0.0780& {\ul 0.0005}	& 0.9525&	0.9961 &{ \ul 0.9992} &	22.5\\\hline
         
        \end{tabular}
     }
    \caption{Depth and normal metrics for several architectures and supervision modes. Best results per dataset are bolfaced, second best underlined.}
    \label{tab:metricsGT}
\end{table*}\begin{table}[h!]
\centering
    \resizebox{\linewidth}{!}{
        \begin{tabular}{clcccc}
        Dataset & Architecture &  Supervision & SSIM $\uparrow$ & MAE $\downarrow$ \\ \hline
        \multirow{1}{*}{Synthetic} 
         & LightDepth U-Net &Light& {0.9901} & {0.0192} \\ \hline
        \multirow{4}{*}{C3VD} 
         &  LightDepth U-Net &  Light & 0.9765 & 0.0657 \\
         & LightDepth DPT &Light& 0.8873	&0.0599&\\
         & LightDepth U-Net & Light (TTR) & \textbf{0.9811} & \textbf{0.0276} \\
         & LightDepth DPT & Light (TTR) & 0.8977	 &  0.0329
         \\
         \hline
         
        \end{tabular}
     }
     \caption{SSIM and MAE for rendered images in C3VD. Test-time refinement (TTR) gives a substantial improvement.}
    \label{tab:metricsGTphoto}
\end{table}\subsection{Datasets}\label{sec:setup} We evaluated LightDepth and relevant baselines on three endoscopy datasets: An in-house \emph{synthetic colon}, \emph{C3VD}~\cite{bobrow2022}, and \emph{EndoMapper}~\cite{azagra2022endomapper}. With these, we can show quantitative and qualitative results with several levels of realism.
\textbf{Synthetic Colon.} 
We simulate a real Olympus CF-H190L endoscope consisting on a fish-eye camera and a spot-light source, both calibrated as in \cite{azagra2022endomapper}. This is in contrast to other synthetic datasets that simulate arbitrary camera and illumination configurations, typically pinhole cameras with no or arbitrary distortion and ideal light sources with no radial falloff. \cite{rau2019implicit, zhang2020template, ozyoruk2021endoslam, rau2022bimodal}, We rendered the images using ray-casting techniques, in which the colon's geometry and albedo are defined by a triangle mesh obtained from a CT scan of a real colon \cite{incetan2021vr}. We ignore global illumination effects and assume Lambertianity, so there are no specular reflections. The influence of these two effects will be assessed in the two other datasets. Our synthetic data is hence composed by 1620 fish-eye RGB frames annotated with per-pixel albedo, depth and normals. We split it into 1168 images for training and 452 images for test. Example frames can be found in the supplementary material.


\textbf{C3VD}~\cite{bobrow2022} contains real images recorded in a phantom with ground-truth depth. The images have been captured by a real Olympus CF-HQ190L endoscope in a phantom silicone model of a human colon. The data is annotated with ground-truth depth and normals by applying 2D-3D registration of the 3D phantom models. The authors claim that the silicone material is opaque, hence we can assume that the only light source available is in the endoscope. Finally, it includes a geometrical calibration based on the Scaramuzza model \cite{scaramuzza2006}. C3VD provides a good compromise between realism (real endoscope, global illumination effects and specular highlights) and ground-truth labels for quantitative evaluation. Of the 10,088 images available, we use 7,200 for training and 2,888 for testing. In the supplementary material, we provide the sections of the phantom used for testing and training.

\textbf{EndoMapper}~\cite{azagra2022endomapper} provides the most challenging data, as it contains real colonoscopy and gastroscopy procedures inside the human body, performed by endoscopists on a day-to-day basis. Here we find real textures such as veins, blood and dirt, and other effects such as blur, water and frames very close or even hitting the mucosa. Foam and bubbles are indeed very common in endoscopy images and are usually ignored. LightDepth is capable of disentangling these as part of the albedo and not of the depth. Before processing the dataset, we perform a manual inspection of the selected sequences and we eliminate occluded and excessively blurred frames.

Finally, we train in three procedures, consisting of two colonoscopies and one gastroscopy. There are a total of 24,444, 23,456 and 3,032 frames, respectively. Details of the sequences and frames we use can be found in the supplementary material.

\subsection{Metrics, Baselines, and Training Details}

We report results using a median-based scale alignment for all methods, even those supervised with real-scale depth, for fairness.
In our experiments, we compare against models that use depth supervision and multi-view self-supervision. For depth supervision, we use two different architectures, U-Net with L1 loss as a representative of convolutional architectures and DPT-Hybrid~\cite{Ranftl_2021_ICCV} as a state-of-the-art representative of transformer-based models, learning inverse depth with an scale invariant loss.

For a fair comparison, we also evaluate our LightDepth using the same U-Net and DPT architectures.
The U-Net is pre-trained on ImageNet dataset \cite{deng2009imagenet}. For DPT, we initialize with the author-provided weights for encoder and depth decoder. The albedo decoder is trained from scratch.
During training, we select a smoothing weight $\lambda_s = 0.1$ in Eq.~\ref{eq:loss} and a learning rate of $10^{-4}$ for the  Adam optimizer. In the synthetic dataset, we trained our network with $\lambda_{sp}$ = 0, as synthetic dataset has no specular reflections. In C3VD and EndoMapper, we use $\lambda_{sp} = 1$.

 \textbf{Test-Time Refinement (TTR)}.  As our LightDepth enables single-view self-supervision, we can continuously refine the depth predictions online, obtaining much more accurate reconstructions. In the results denoted as ``(TTR)'', we perform online test-time refinement for each test image separately during $N = 20$ optimization steps, using the loss $\mathcal{L}$ in Equation \ref{eq:loss}, as in training time. To mitigate the risk of catastrophic forgetting, we load again the original model trained in the train split after TTR for each image.
 
\begin{figure}
  \includegraphics[width=\linewidth]{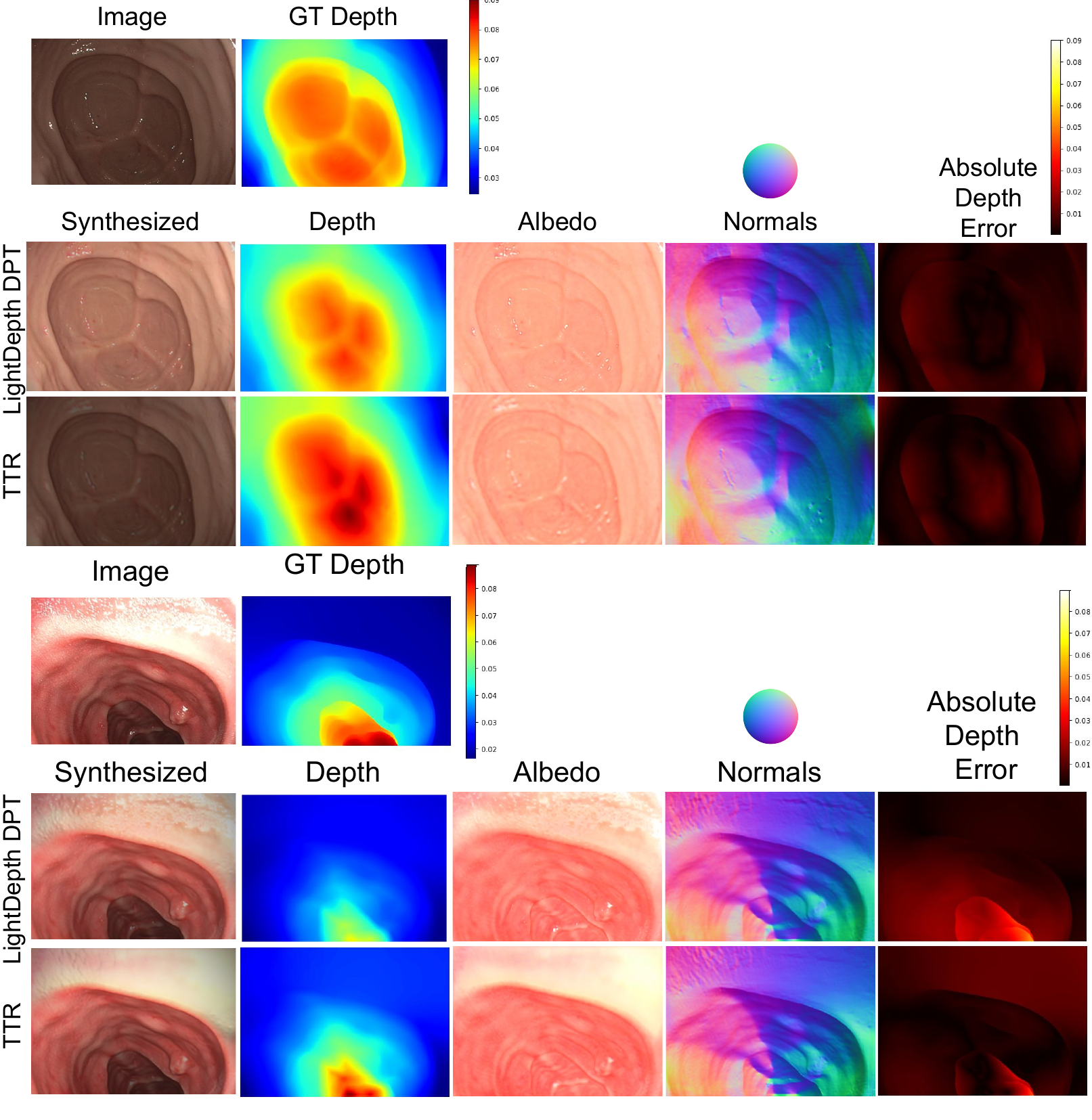}
\caption{DepthLight and DepthLight TTR on C3VD. Our light decline captures the correct shape of the cecum in the first image and the shape of the polyp in the second. Note how the estimates of normals and albedo are similar before and after TTR. By optimising depth by reducing illumination, DepthLight achieves a darker appearance and improvements in depth estimation.}
\label{fig:TTR}
\end{figure}

Note in Table~\ref{tab:metricsGT} how TTR improves significantly the metrics with respect to LightDepth without TTR for U-Net and DPT architectures. Remarkably, observe how TTR even outperforms the metrics achieved by Depth GT supervision. Figure \ref{fig:TTR} shows the improvement given by TTR in the network prediction of depth, normals and albedo and overall in the 3D reconstruction.
Inference time is $\sim 5$ms for LightDepth U-Net and $\sim 22$~ms for LightDepth DPT on a NVIDIA GeForce RTX 3090. We can do TTR in $\sim 90$~ms per optimization step in U-Net and $\sim 190$~ms in DPT.




\subsection{Quantitative Results on Synthetic and Phantom}
\label{sec:quant}

\paragraph{Synthetic colon.}

The first two rows in Table~\ref{tab:metricsGT} report depth and normal metrics for a U-Net supervised with Depth GT, and our self-supervised LightDepth U-Net architecture. Observe that the metrics are similar. This is notable, as self-supervision is consistently reported in the literature to underperform with respect to depth supervision, and suggests that illumination decline provides a very strong self-supervisory signal in endoscopies, which our experiments in the other two datasets confirm.

Furthermore, light self-supervision outperforms Depth GT supervision in MedAE and $\delta < 1.25$, which means that most of the error distribution is lower for light self-supervision and only a small fraction of large errors are better with depth supervision. We observed that it is in far and dark areas where light self-supervision is weaker and this produces a higher depth MAE and RMSE. Observe the significantly lower  error in normal with our light self-supervision, due to the lower errors in most pixels.

\paragraph{C3VD Phantom.}

We report depth and normal metrics on the real phantom images of the C3VD dataset in Table~\ref{tab:metricsGT}. Our self-supervised architectures LightDepth U-Net and LightDepth DPT with TTR outperform supervision with Depth GT in MedAE, while the rest of the metrics are very close. As in the case of the synthetic dataset, this is a remarkable result because self-supervised architectures typically lag behind supervised ones in single view depth estimation. 
The fact that LightDepth MedAE is better and RMSE is worse suggests that our errors are better in most of the distribution, and there are a few regions with large errors where Depth GT supervision is able to offer an advantage. Table \ref{tab:metricsGTphoto} details metrics on the quality of the rendered image, which suggest the strength of the self-supervision signal. Observe the improvement of this metrics for the TTR case. 

In Table \ref{tab:metricsGT}, observe that the multi-view self-supervised baselines, Monodepth2~\cite{monodepth2} and CADepth~\cite{yan2021channel}, have a poor performance in our data, worse in comparison than results in other datasets. This could be due to the weak textures and changing lighting in the colonoscopy images, resulting in noisy estimations for relative motion and uninformative photometric residuals. Being single-image, our approach is impervious to such difficulties.

\textbf{Domain shift}. As synthetic-to-real is common in endoscopies to address the lack of ground-truth depth for supervision, we also evaluated XDCycleGAN \cite{mathew2020augmenting} as a baseline. Note that the domain shift is still affecting the results. Our single-view LightDepth self-supervision enables training in the target domain, and hence removes completely the domain shift, achangedchieving significantly lower errors. 

\begin{table}[h!]
\centering
\resizebox{\linewidth}{!}{%
\begin{tabular}{cc|l|ccc|c}
 \multicolumn{2}{c}{Dataset}            & & \multicolumn{3}{c}{Depth [mm]}                           & Normals [$^\circ$]    \\
Train                  & Test                   & {Supervision} & MAE  $\downarrow$ & MedAE $\downarrow$ & RMSE$\downarrow$ & MAE $\downarrow$ \\ \hline
\multirow{2}{*}{Synt.} & \multirow{2}{*}{Synt.} & {Depth GT}       & 4.37    & 2.99     & 6.38    & 25.1   \\
                       &                        & {Light}       & 4.76    & 2.47      & 8.60    & 15.2    \\\hline
\multirow{4}{*}{Synt.} & \multirow{4}{*}{C3VD}  & {Depth GT}       &  9.44            & 5.79               & 12.83            & 73.7             \\ 
                       &                        & {Light}        & 5.09            & 3.51      & 7.14    & 27.7   \\ 
                       &                        & {Depth GT (TTR)}  & 4.96            & 3.14 & 7.11 & 25.4 \\
                       &                        & {Light (TTR)}  & {\ul 3.80}           & \textbf{2.51} & 5.54 & { \ul 23.6} \\\hline
\multirow{3}{*}{C3VD}  & \multirow{3}{*}{C3VD}  & {Depth GT}        & { 4.15}    & 3.29      & {\ul 5.52}    & 26.5 \\
                       &                        & {Light}        & 4.37    & { 2.92}      & 6.31    & 24.0 \\
                       &                        & {Light (TTR)}        & \textbf{3.72}    & { \ul 2.59}      & \textbf{5.43}    & \textbf{23.5} \\\hline

\end{tabular}%
}
\caption{Synthetic-to-real domain shift. Best results in C3VD test set are boldfaced, second best are underlined. Note the domain shift effect between Synt. and C3VD test data in the bigger errors, and how TTR removes the domain shift effect completely. Notably, our LightDepth TTR delivers similar errors than the models without domain shift, trained in C3VD.}
\label{tab:transfer}
\end{table}

Table \ref{tab:transfer} elaborates further on domain shift by showing depth and normals metrics for a U-Net architecture in these cases. Specifically, we trained a U-Net model with Depth GT supervision and light self-supervision in our synthetic dataset and evaluated their performance in the synthetic and C3VD test sets. Observe how the domain shift affects all metrics significantly. Interestingly, the model trained with light self-supervision and without TTR generalizes significantly better to the C3VD data, as our LightDepth self-supervised model is closer to the physical phenomena than Depth GT supervision. Again, note that single-view self-supervision removes completely the domain shift effect, as models can be trained directly in the target domain. Very remarkably, the performance of our models with domain shift after TTR matches the performance of the models without domain shift.

\begin{wraptable}{r}{3.0cm}
\centering
\scriptsize
\begin{tabular}{c  c} 
 Method  & MAE [$^\circ$] \\
 \hline
 U-Net& 16.24\\
 TFtN~\cite{fan2021three}	& 3.89\\
 Open3D~\cite{Zhou2018Open3D} & 1.67 \\
 In-house & \textbf{1.32}\\
 \hline
\end{tabular}
\caption{Normal's MAE for baseline methods.} 
\label{tab:ablationNormals}
\end{wraptable}
\textbf{Normals from Depth}. The literature details different manners to obtain surface normals from a depth map, e.g.,~ \cite{fan2021three,boulch2016deep}. Table \ref{tab:ablationNormals} shows a MAE analysis of the most promising ones in C3VD. Specifically, we evaluate four methods: a U-net trained to regress normals from depth, the recent TFtN method~\cite{fan2021three}, the implementation in Open3D~\cite{Zhou2018Open3D} that computes normals from a k-nearest neighbourhood in the point cloud, and an in-house method that uses six-neighbourhood in the image. Our analysis shows that an analytic average in a neighbourhood is significantly better than a U-Net and TFtN, and our in-house method that considers a neighbourhood in the image is slightly better, so this last one was our choice.


\textbf{Ablation Study on the Loss}. In Table~\ref{tab:loss_ablation}, we ablate the terms of our loss function. The smoothness prior ($\mathcal{L}_s$ term) is remarkably beneficial for both depth and normal prediction. When we do not take advantage of the information of the specular reflections (no $\mathcal{L}_{sp}$ term), we obtain worse results. Adding this new loss term, we see how all the depth and normal metrics improve, especially in the median error, which outperforms the supervised and now matches that obtained in the simulation experiment. Still, the depth MAE and RMSE are slightly higher than those of the baseline due to the far spurious points.

\begin{table}[h!]
\centering
\resizebox{\linewidth}{!}{%
\begin{tabular}{cccccc}
 & \multicolumn{3}{c}{Depth [mm]} & Color & Normals [$^\circ$]    \\
Loss & MAE  $\downarrow$ & MedAE $\downarrow$ & RMSE$\downarrow$ & MAE $\downarrow$ & MAE $\downarrow$\\ \hline
$\mathcal{L}_p$ & 6.05 & 3.93  & 8.79    & 0.0637    & 35.5  \\
$\mathcal{L}_p + \mathcal{L}_s$ & 4.95 & 3.04 & 7.23  & 0.0690 &   24.6  \\
$\mathcal{L}_p + \mathcal{L}_s + \mathcal{L}_{sp}$ & 4.37 &  2.92      & 6.31     & 0.0657  &  24.0 \\
                     \hline
\end{tabular}%
}
\caption{Ablation study of the losses with LightDepth U-Net in C3VD dataset. Observe the improvement given by each term.}
\label{tab:loss_ablation}
\end{table}
\begin{figure*}
  \includegraphics[width=\textwidth]{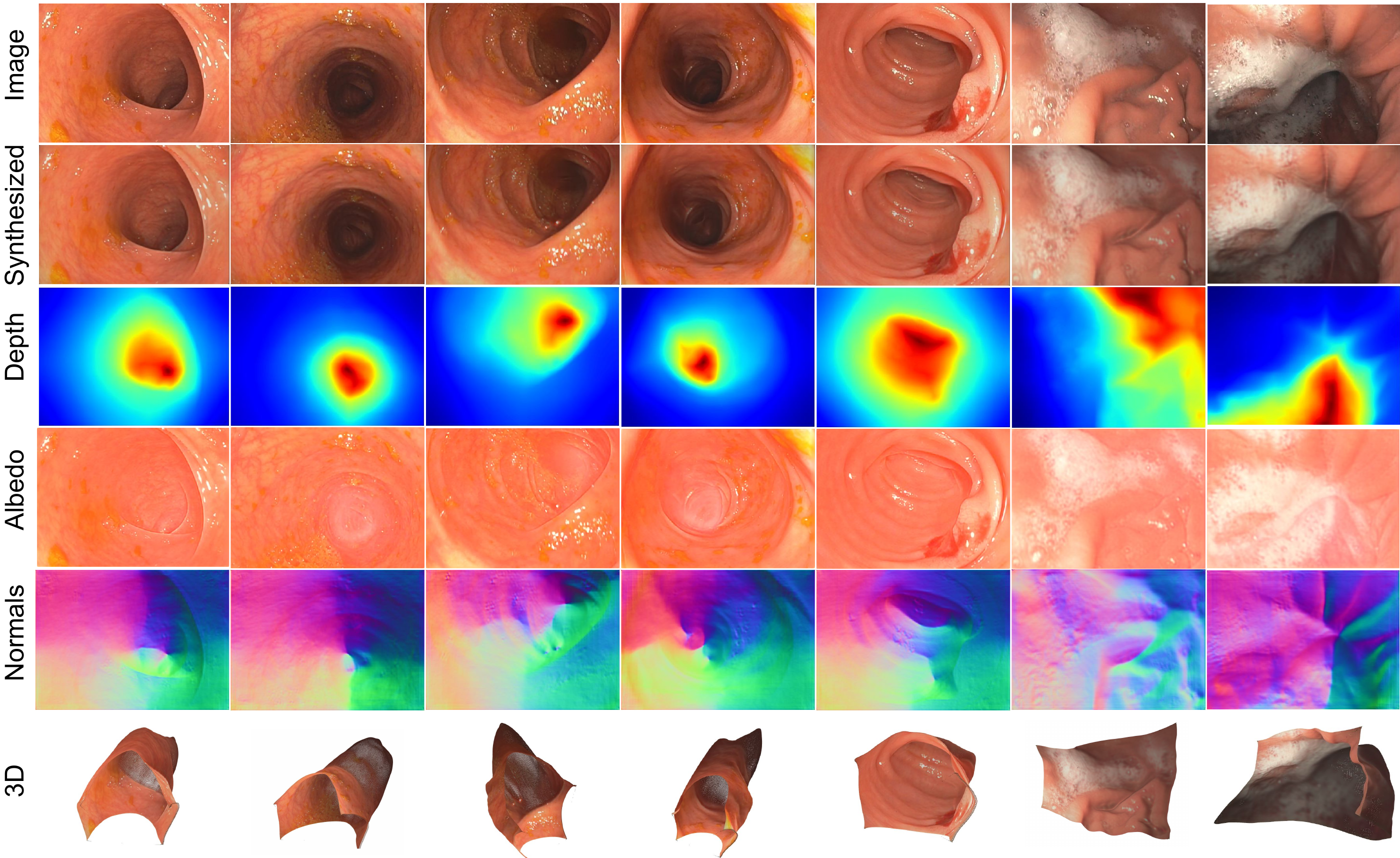}
\caption{Qualitative results on EndoMapper with LightDepth DPT. Columns 1--5 are real colonoscopy images, and columns 6--7 are real gastroscopy images. In colonoscopies, observe that the normals exhibit a tubular shape specific of the colon. The albedo prediction captures disruptions such as veins, blood, dirt, foam and specularites. Note the influence of light decline in the image and the correlation with the estimated depths.}
\label{fig:all}
\end{figure*}

\subsection{Qualitative Results in Real Endoscopy}

We now turn to real images of a human colon from the EndoMapper dataset and present qualitative results in Figure \ref{fig:all}. Additional ones can be found in the supplementary material. Some details are recovered very accurately, such as the normal maps showing clearly the tubular shape; the depth maps reflecting the discontinuities in the Haustras; the albedos capturing the blood vessels, in particular in the $\text{5}^\text{th}$ column;  and the bubbles and fluids colors in the $\text{6}^\text{th}$ and $\text{7}^\text{th}$ columns, which make the 3d reconstruction of these bubbles and fluids very plausible. 

Unfortunately, there is no ground-truth data available for this dataset, which prevent us from presenting quantitative results, and we do not know of any other dataset with real colonoscopy images that includes ground-truth data. Nevertheless, visual inspection of our results hints that the strengths of our techniques demonstrated quantitatively in Section~\ref{sec:quant} will carry over on truly realistic scenarios like this one. 
\section{Limitations and Discussion}
As mentioned in Section \ref{sec:losses}, our depth predictions are up-to-scale. Even if the camera auto-gain was available, the albedo scale may be challenging to learn, so estimating the real scale is not straightforward. In any case, other methods such as multi-view self-supervision or synthetic-to-real cannot guarantee an accurate estimation of the scale either.
We assume that Lambertian reflectance is prevalent in most tissues, and for areas where this does not hold, we use a basic model to capture specularities. Further research could focus on the application of more sophisticated photometric models that cover specularities, e.g., the Phong model.

Thanks to our priors on albedo and depth, we successfully disentangle both factors in our experiments. However, our $V=100$ prior might not hold in areas of clotted blood or with very dark albedos, e.g., because of a disease. These priors might need to be tuned in new application domains for enhanced performance.
Finally, although we demonstrate this technology in the context of endoscopy, its principles are applicable in any setup in which the only light source is close to the target surface and rigidly attached to the camera. In other words, our LightDepth has the potential to open research avenues in many other domains.
\section{Conclusions}In this work, we have proposed, for the first time, a single-view self-supervision method for depth learning, which we denote LightDepth, that exploits and is limited to the case of a single spotlight source co-located with a monocular camera, a case that includes, among others, the relevant application of medical endoscopy. As our main contribution, we developed the specific self-supervised learning setup that models the quadratic light decline and enables self-supervised learning. We have implemented two different architectures, a first one based on convolutions and a second one based on transformers, and evaluated their performance against ground-truth supervision, multi-view self-supervision, and domain transfer approaches. Our results show that LightDepth outperforms multi-view self-supervision and synthetic-to-real transfer and matches the performance of fully supervised approaches. Not only that, its training and test-time refinement setup is significantly simpler: LightDepth only requires a reasonable endoscope calibration and does \emph{not} require camera motion estimation nor ground-truth labels nor realistic simulations, all of them challenging in endoscopies. This unlocks, from a practical point of view, relevant potential applications in the medical domain.

\section*{Acknowledgements}
This work was supported by the EU Comission (EU-H2020 EndoMapper GA863146), the Spanish Government (PID2021-127685NB-I00, FPU20/0678, PID2021-125209OB-I00, PGC2018-096367-B-I00 and TED2021-131150BI00) and the Aragon Government (DGA-T45\textunderscore23R).
{\small
\bibliographystyle{ieee_fullname}
\bibliography{egbib}
}

\end{document}